\documentclass[a4paper,11pt]{article}
\usepackage[utf8]{inputenc}
\usepackage{bm,amsmath,verbatim}
\usepackage[round,sort,comma]{natbib}

\title{What are Neural Networks made of?}
\author{Rene Schaub \\ { \texttt{renecschaub@gmail.com}} }
\date{}

\begin{document}

\maketitle

\begin{abstract}
\normalsize
The success of Deep Learning methods is not well understood, though various attempts at explaining it have been made, typically centered on properties of stochastic gradient descent. Even less clear is why certain neural network architectures perform better than others. We provide a potential opening with the hypothesis that neural network training is a form of Genetic Programming.
\end{abstract}

\section{Introduction}
A long-standing question in machine learning has been the perplexing effectiveness of Deep Learning neural networks (NN) in solving various prediction problems. The ingredients that make good neural networks are known, with results surpassing human cognitive abilities in some restricted tasks; but we are still not sure why they work so well, or at all, for that matter. It is difficult to understand how a huge non-convex optimization problem that is NN can be solved well enough by simple stochastic gradient descent (sgd).
\citeauthor{tishbyBlackBox} presented evidence that neural networks undergo a compression phase during training, with successive layers $T$ retaining less and less mutual information with the input variable $I(X;T)$, while building up mutual information with the target variable $I(T;Y)$, compatible with the \textit{information bottleneck} \citep{bottleneck}
$$\min_{p(t|x)} I(X;T) - \beta I(T;Y)$$
The empirical evidence appears to be strong, and we have not seen any convincing arguments to the contrary \citep[e.g.][]{stupidMit}. Effective learning, as represented by Deep Learning, would appear to necessitate that only relevant information be retained \citep[see also][]{uclaDrop}, to approach the structure of the target variable. As shown in \citep{tishbyError}, quantizing, or similarly, clustering \citep{stupidMit} or saturating variables on intermediate layers may be required to attain good generalization from a finite number of input samples. Such a process would by necessity reduce $I(X;T)$, satisfying the information bottleneck.
NN activation functions have saturation regions (Sigmoid, ReLu), so this is certainly plausible.
But all this is still not addressing the core problem of local minima in global optimization. Somehow, through extreme redundancy of parameters in NN, there are supposed to be so many possible good solutions, that, with the help of a little noise, a gradient path can plow through to one of them. 
While interesting properties of the diffusion process initiated by sgd have been found \citep[e.g.][]{uclaDrop}, nothing clearly ties to the observed information bottleneck.

\medskip

As any practitioner may observe, during the training process, neural networks find simple solutions first and then go on to more complexity. For an example in language modeling, BERT \citep{bert} will first attain the trivial solution, which is the frequency solution. It will predict a masked word as the single word that is correct most often when plugged in, which in many languages and with the given tokenization is the comma ','. From analysis of convolutional neural networks (CNN) we know that lower layers represent basic local features, and higher layers more abstract concepts related to the classification problem \citep[see e.g.][]{cnnAbstract}. Deep Learning apparently builds its solutions as a sequence and hierarchy of concepts, which has been suggested to also be taking place in human learning \citep{tenenbaumCompositional}.

\subsection{Preliminaries}

We only consider feed-forward neural networks in this presentation. We interchangeably make use of the terms \textit{features, programs}, and \textit{tensors}, to highlight specific aspects, or simplify the exposition. We also refer to \textit{neurons} and \textit{nodes} interchangeably. When refering to \textit{tensor} or \textit{subtensor}, we mean a  multidimensional array section of the output neurons of a NN layer. The corresponding program is taken to be the tree - or rather, directed acyclic graph (DAG) - of operations and sample inputs from which the tensor is computed in a \textit{forward pass}, which is the evaluation of all nodes in a neural network for a given batch of samples. A feature can be a single neuron of a layer, or a tensor of neurons. It is convenient to sometimes consider a feature to be a binary indicator of a signal, which can be active or inactive, depending on the input to the neural network, or slightly more generally, a feature may be a scalar correlate as used in regression analysis. 

\section{NN as Feature Regression} \label{regression}
\def\pc(#1, #2){p\left( #1 \middle\vert #2 \right) }
\def\qc(#1, #2){q\left( #1 \middle\vert #2 \right) }

Cutting the directed acyclic computation graph that makes up a maximum likelihood, feed-forward neural network into two parts, we can split the NN into features obtained from the lower part of the cut, and a distribution $q$ parameterized by $\tau$ consisting of the top part of the graph, taking as its input the features.
As illustrated in \citep{selection}, the conditional maximum likelihood loss function $l$, here for illustration represented as the limit $n \to \infty$ of the number of samples $n$, breaks down into the following terms, given a selection of features $\theta$. The remaining, unused features are indicated by $\tilde{\theta}$, so that the predictor variables $X = \{X_{\vphantom{\tilde{\theta}}\theta}, X_{\tilde{\theta}}\}$. $p$ stands for the true conditional distributions of the target variable $Y$.

\begin{equation} \label{split}
 \begin{split}
  l & = - E_{\bm{x}y}  \log \qc(y, {\bm{x}_\theta,\tau}) \\
    & = E_{\bm{x}y}  \log \frac{\pc(y, {\bm{x}_\theta})}{\qc(y, {\bm{x}_\theta,\tau})}  
    + \log \frac {\pc( y, {\bm{x}} )}  {\pc( y, {\bm{x}_\theta} )} 
    - \log \pc( y, {\bm{x}} ) \\
    & =  E_{\bm{x}y}  \log \frac{\pc(y, {\bm{x}_\theta})}{\qc(y, {\bm{x}_\theta,\tau})}  
    + \log \frac {\pc( y\bm{x}_{\tilde{\theta}}, {\bm{x}_{\vphantom{\tilde{\theta}}\theta}} )}  {\pc( y, {\bm{x}_\theta} ) \pc( \bm{x}_{\tilde{\theta}}, {\bm{x}_{\vphantom{\tilde{\theta}}\theta}} )} 
    - \log \pc( y, {\bm{x}} ) \\
    & =  E_{\bm{x}y}  \log \frac{\pc(y, {\bm{x}_\theta})}{\qc(y, {\bm{x}_\theta,\tau})}
    +  I( Y; X_{\tilde{\theta}} \vert X_{\vphantom{\tilde{\theta}}\theta} )
    +  H(Y \vert X) \\
    & =  E_{\bm{x}y}  \log \frac{\pc(y, {\bm{x}_\theta})}{\qc(y, {\bm{x}_\theta,\tau})}
    +  I( Y; X) - I( Y; X_{{\theta}} )
    +  H(Y \vert X)
 \end{split}
\end{equation}

The last equality follows from the chain rule for joint probabilities, which carries over to mutual information $I$.
With powerful $q$ - NN are universal function approximators, even with usual width of layers \citep{universalWidth} - there is no reason to assume a trade-off between how well $q$ can approximate $\pc(y, {\bm{x}_\theta}) $ depending on which features are provided. Although it clearly is much more difficult to use features $\theta$ with high $I(Y;X_\theta)$ but complex structure unrelated to $Y$, than to use simple correlates of $Y$ that may have lower mutual information.  Gradient descent may converge to a local optimum that is not close to the global optimum. Sgd noise and dropout noise may help somewhat in this case by 'jumping' out of some local minima. But the optimal solution space may not be reachable on any path close to the gradient. The split of the ML objective in (\ref{split}) into a Kullback-Leibler Divergence term and a feature selection component allows us to illustrate a \textit{sequential} process of adding features - which is what we surmise takes place during NN optimization by sgd. Adding a feature can improve the objective by up to its mutual information with the target, in case it is independent of the existing features, and if the updated KL term can be successfully optimized. \citep[Corollary 13]{selection} shows it is also possible that adding a given feature to $\theta$ would have no impact at one time, but could improve the conditional likelihood once other features had been added first to $\theta$. An excluded feature may therefore become useful in the future.

\section{Overview of Genetic Programming}

We will provide a quick overview of Genetic Programming (GP) \citet{gp}, and make superficial comparisons of these concepts to NN, when the connection is obvious.

GP is a very compute intensive, but general, method for generating a computer program that can be evaluated to have a high \textit{fitness} score, and that takes inputs and generates outputs. Note that this format is also shared by NN or statistical regression methods, except that evaluation is restricted to independent samples and the program to continuous functions. Before deep learning NN achieved breakthrough results in some areas that heretofore were only achievable by human intelligence or cognition, GP had shared a similar  distinction of creating, for example, designs that were competitive with human inventions \citep{patents}. It is noteworthy that while GP creates many useless programs during its optimization run, and thus is often perceived as inefficient, these results were obtained over a decade ago, on moderate hardware by today's standards.

\subsubsection{Programs}
A program is a tree of nested primitive function call instances. Each function comes from the same small set of basic functions such as '+', 'OR', 'IF', or an input node itself, and can have multiple arguments. It basically is a functional program that can be evaluated by traversing the tree. 
In comparison, every neuron in a neural network is computed from a typically highly regular DAG of basic (sub)differentiable function calls, such as ReLu activations or linear maps. It also is evaluated by traversing the DAG. So NN neurons or tensors are comparable to GP programs. 

\subsubsection{Candidates}
Unlike numerical optimization, which only traces one point in the parameter space, GP maintains a large set of intermediate diverse programs during each phase, the current \textit{generation} of \textit{candidates}, and discards candidates that don't advance to the next generation.
Instead of looking at the parameters in NN as one space, we can see each layer as set of features, or programs. Higher layers are constructed from lower layers, in a reusable fashion. While not going into any details, program reuse is also a feature of GP. 


\subsubsection{Random Combinations}
In each new generation, GP probabilistically retains candidates according to fitness, and in addition creates new candidates from combinations of random pairs, in a genetically inspired manner. Each pair exchanges two randomly chosen subtrees. This so-called \textit{crossover} operation is the main driver for generating new programs.

\subsubsection{Initialization}
GP generates initial programs by randomly creating trees of various sizes. Most of these trees have very poor fitness, yet typically are sufficient to jump start the process.
NN parameters are always initialized randomly, typically as Gaussians. Initialization is extremely unlikely to generate higher-level features at the start, but some small, weak, but usable basic features are likely to be created in this unguided fashion.

\subsubsection{Optimizaton Process}
Each iteration produces a new generation from combinations of the previous generation of programs, until fitness of the best program in a generation is deemed good enough.

\section{Evaluation and Selection by Gradient} \label{selection}
An essential step in GP is the evaluation of programs for fitness, and the following selection of candidates for the next generation.
Since NN parameters are randomly initialized, on average each randomly constructed feature will at first have similar strength along its path(s) to the top layer, and similar strength in the paths that make up its program tree, down to the input nodes (though some weights will be characteristic of the feature). The path to the objective may be very short, as is the case when skip links are employed. Assuming active paths exist, we may, for instance, consider $q$ to be the final linear combination and softmax that form the probability distribution for maximum likelihood. Just as in multinomial logistic regression, maximum log likelihood with likelihood $q$ is convex, and thus eminently solvable to the global optimum with plain gradient descent, so from this standpoint, it is no mystery why NN is able to find a solution, given the features. The gradient $\nabla{l}$ will increase weights for features that are correlated with the target, but have not been weighted sufficiently yet. One or multiple gradient steps therefore amount to selection of features, by reinforcing their paths. This follows from the calculus chain rule. For example, branches in a computation tree are typically induced by the '+' operator. In neural networks the chained gradient splits along these branches into paths, and any multiplicative parameters along a path to a feature contribute to its strengthening. Note that the issue of the \textit{vanishing gradient} in deeply layered networks has been solved, through the use of, for example, batch normalization \citep{batchNormalization} or layer normalization \citep{layerNormalization}.

Evaluation of features as in GP is implicit in the evaluation of the gradient, and a step in approximately the direction indicated by the gradient amounts to selecting and reinforcing of features. 

After sufficient strengthening of useful features, they are now in a position to become components of higher-level features: Random combinations that before were just noise, now are new candidates for the following iterations. More accurately, the random combinations will have changed, because the underlying feature weights, which are inputs to the random map resulting in the combination, have changed, and with sufficient strength of the component features, the resulting combination will register to the gradient above the noise threshold.
We propose that gradient steps used in optimizing neural networks implicitly perform the GP steps of evaluation and selection of features, preparing the ground for random combination of these features as new candidates.

Only features on lower layers can be combined to produce more advanced programs on higher layers. Initially, a feature may be redundantly represented, by chance, on many layers, which assists convergence. As time goes on and the interchangeable features have been strengthened, fewer of them may be required for the same effect, so that only instances on lower layers persist, because they are needed as part of higher-level features. This dynamic would account for the observation \citep{cnnAbstract} that layers in convolutional neural networks (CNN) represent progressively more abstract concepts.

\section{Pairing of Candidates}
To uncover another operation of GP represented in NN, and to make the point that better neural network architectures are also more authentic GP implementations, we now turn to the recent neural network architecture that is easily the most successful as of 2019, BERT \citep{bert}. 
Only a couple of years ago, the prevalent opinion on the capabilities of Deep Learning could have been summed up in a quote by Andrew Ng in 2016:
\begin{quote}
 If a typical person can do a task in less than 1 second, we can automate it with AI.
\end{quote}
The flip side of this statement suggests that DL may only be effective at very short perception tasks. With BERT, it is clear now that yard stick has to be put much farther. The language models learned by BERT are competitive, or almost competitive, with human comprehension of several paragraphs of text.

BERT at its core is a denoising autoencoder \citep{daeDensity, daeGeneral} with  its self-supervised marginal conditional maximum likelihood objective and noisy input, coupled with the internal architecture of the Transformer \citep{allYouNeed} without masking, therefore bi-directional.

The essence of the Transformer is summarized by the following operations on the key tensor $k$, question tensor $q$, and value tensor $v$ from the previous layer, using the same names as in \citep{allYouNeed}.

\begin{align}
& s_{i j} = \sum_w q_{i w} k_{j w} \\
& s'_{ij} =  \frac{ e^{s_{ij}} } { \sum_{j'} e^{s_{ij'}} }  \label{softmax} \\  
& v'_{iz} = \sum_j s_{ij} v_{jz}   + v_{iz}
\end{align}

The new value tensor $v'_i$ at position $i$ is the sum of the old value tensor $v_i$ and effectively, value tensor $v_j^*$ at position $j^*$ where the selector $s_i'$ obtained from the softmax operation (\ref{softmax}) is peaking.

One of the key operations of GP is combining random \textit{pairs} of existing programs to form new candidates. Looking at the Transformer from this angle, this is exactly what it does. A subtensor from location $j^*$ is added, via the skip link, to a subtensor at location $i$, at once giving the pairing of $i,j^*$, and its combination.

It is not hard to argue that this selection is random, as required by GP; actually it would be much harder to argue that it is mediated by the gradient, which would amount to making discrete choices over multiple local optima. All question tensors and key tensors are randomly mapped (for each unique input sample), from the random initialization of parameters of the linear layers; so their inner products and therefore selection of $j^*$ are random too. 

As an arbitrary structure can be encoded in a subtensor, simple addition can correspond to an arbitrary combination, approximating the \textit{crossover} operation employed by GP. Also, skip links can connect subtrees from any layer, further strengthening the analogy. That being said, the crossover combination is not the last word on ideal combinations in GP, and many others may do as well. Key for GP is the random, \textit{pairwise} combination of fit programs.
Therefore we argue that the Transformer's essential function is to randomly pair up and combine successful features, just as GP does.

\section{Sources for Random Combinations}
We hypothesize that random initialization of NN parameters serves as a main store of random numbers for generating random combinations of features during the training process. 
It is obvious that at the start of gradient descent, the pairing of locations will be at random due to this initialization. It is critical that any useful random combination be maintained across iterations, to preserve the successful feature. Initialization of parameters only takes place once in the beginning, so parameter values are generally maintained except as updated by the gradient step and the sgd diffusion process. A successful feature will be reinforced along its paths by the gradient.

As gradient descent progresses, random parameters in unused locations continue to serve in presenting further random combinations, as argued in section \ref{regression}, now of progressively deeper combinations of features, analogous to GP. Of course it is essential that the NN be highly redundant and deeply parameterized. When this is not so, it can be expected that a single NN run may not necessarily converge to an acceptable solution. This is because not enough random combinations and candidates are generated to be effective for GP. It used to be common practice to perform multiple NN runs during an extensive hyperparameter search. Each run of course receives a unique random instantiation of parameters.

But here is where the sgd diffusion process provides additional randomness for feature combinations, to supplement the reservoir of parameter initializations
(BERT has over 300 million random parameters, mostly packed into its \textit{expansion} layers). The random error in the stochastic gradient is added to NN parameters $\tau$ in every step $\tau' = \tau + \alpha (\nabla l + \epsilon)$. This results in a random walk, which is a strongly autocorrelated process. The values it generates can be expected to change slowly, allowing gradient steps to solidify features before they disappear again (and protect utilized features from erosion by the diffusion process). 

Let's look at some of the numbers in the BERT model, to see if the potential feature candidate population is large enough to support GP. The characteristic pairing operation is over the dimension of 512 input positions. This does not mean that the candidate population is only 512 candidates large. Each position consists of 16 \textit{attention heads} (independent random pairings), and a \textit{depth} vector of 1024 neurons. If we take each neuron to be a potential feature, this presents at every layer a candidate population of over 8 million.

\section{Discussion}
We have presented somwhat plausible arguments that the mechanics of recent neural network architectures and principles support the operations characteristic of Genetic Programming. These include the random pairwise combinations of fit programs to produce new candidates. It is thus possible that the power of neural networks derives from evolutionary search, using fast parallel gradient-based evaluation and selection. New NN architectures could be guided by how well they enable GP.

While various links have been proposed between human brain function and neural networks \citep[see e.g.][]{tenenbaumBrainMachine}, a connection to evolutionary computation would suggest itself quite naturally. 

\bibliographystyle{plainnat}
\bibliography{paper}

\end{document}